\documentclass[letterpaper, 10 pt, conference]{ieeeconf}  

\IEEEoverridecommandlockouts                              

\usepackage{subcaption}
\usepackage{dblfloatfix}
\usepackage{float}
\usepackage{graphicx}

\title{\LARGE \bf
Quantitative Data Analysis: CRASAR Small Unmanned Aerial Systems at Hurricane Ian
}

\author{Thomas Manzini$^{1}$, Robin Murphy$^{1}$, and David Merrick$^{2}$
\thanks{$^{1}$ Department of Computer Science and Engineering,
        Texas A\&M University,
        College Station, TX 77843, USA
        {\tt\small tmanzini@tamu.edu, robin.r.murphy@tamu.edu}}%
\thanks{$^{2}$Center for Disaster Risk Policy at Florida State University,  Tallahassee, FL 32306, USA
        {\tt\small dmerrick@em.fsu.edu}}%
}

\begin{document}

\maketitle
\thispagestyle{empty}
\pagestyle{empty}

\begin{abstract}

This paper provides a summary of the 281 sorties that were flown by the 10 different models of small unmanned aerial systems (sUAS) at Hurricane Ian, and the failures made in the field.
These 281 sorties, supporting 44 missions, represents the largest use of sUAS in a disaster to date (previously Hurricane Florence with 260 sorties).
The sUAS operations at Hurricane Ian differ slightly from prior operations as they included the first documented uses of drones performing interior search for victims, and the first use of a VTOL fixed wing aircraft during a large scale disaster.
However, there are substantive similarities to prior drone operations. 
Most notably, rotorcraft continue to perform the vast majority of flights, wireless data transmission capacity continues to be a limitation, and the lack of centralized control for unmanned and manned aerial systems continues to cause operational friction.
This work continues by documenting the failures, both human and technological made in the field and concludes with a discussion summarizing potential areas for further work to improve sUAS response to large scale disasters.

\end{abstract}

\section{Introduction}
\label{sec:intro}
Hurricane Ian was a Category 4 Hurricane that made landfall along the central west coast of Florida on September 28, 2022. Hurricane Ian is among the most intense weather events to hit the United States, resulting in at least 156 fatalities and an estimated \$112.9 billion in damages \cite{bucci2022national}.

Hurricanes are unique in the amount and type of damage they inflict and the large area of coverage. 
In terms of the use of sUAS, hurricanes are challenging because of scope. The extent of the disaster  may span multiple geographical and administrative regions. The scale stretches resources and confounds a comprehensive understanding of the entire event. Portions of the affected area are often rural, which may be beyond the range of urban-based helicopters and other crewed aviation assets. Finally, hurricanes are also physically and cognitively demanding on sUAS and operators because the missions are off-normal \cite{murphy2020humans}. 

As it became clear that Hurricane Ian was going to make landfall in the United States, The Center for Disaster Risk Policy at Florida State University organized and led a multiagency sUAS and remote sensing team. 
This team, which was operationally referred to as FL-UAS1, consisted of ten governmental agencies (Alachua County Fire Rescue, Boone County Fire
Protection District (Missouri), The Civil Air Patrol, Florida Department of Law
Enforcement, Jacksonville Sheriff’s Office, Leon County
Sheriff’s Office, Miami-Dade Fire Rescue, Okaloosa County
Sheriff’s Office, Tallahassee Fire Department, Tallahassee
Police Department), two academic institutions (FSU, Texas A\&M), and
one insurance company (USAA).
This team contained sUAS squads capable of flying missions in support of the response operations, and also data managers whose role was to process, organize, sanitize and publish the data collected in the field.

In total the members of FL-UAS1 conducted 281 sorties (flights) and collected 636GB\footnote{This includes one instance where a squad accidentally duplicated their data. We present this number because data managers were not aware of this duplication during the event and so it more accurately characterizes the volume of data that was processed and managed in the field.} in support of 44 missions using 10 different sUAS models. This represents the largest use of sUAS, in terms of sorties, compared to previously documented disasters\footnote{Differences in terminology between ``sorties", ``flights" and ``missions" between different operations leaves some ambiguity in comparisons.}. The details of this comparison can be found in Table \ref{tab:drone_use}. 

\begin{table}[]
\centering
\begin{tabular}{|l|c|c|}
\hline
\textbf{Event}     & \multicolumn{1}{l|}{\textbf{Sorties}} & \multicolumn{1}{l|}{\textbf{Missions}} \\ \hline
Hurricane Ian      & 281                                   & 44                                     \\ \hline
Hurricane Harvey \cite{suasharveyirma, fernandes2018quantitative}  & 112                                   & 56                                     \\ \hline
Hurricane Michael \cite{fernandes2019quantitative}  & 80                                    & 26                                     \\ \hline
Hurricane Florence \cite{florence_use} & $\geq$260                             & $\geq$260                              \\ \hline
Hurricane Irma \cite{suasharveyirma}    & 247                                   & $\geq$500                              \\ \hline
\end{tabular}
\caption{Comparison of Hurricane Ian to prior documented large scale disaster sUAS operations.}
\label{tab:drone_use}
\end{table}


This paper presents summary statistics from the sUAS response to Hurricane Ian, details the mission types and use cases for these sUAS systems during this disaster, discusses failures of both the technological and human systems, and concludes with calls for future work and improvements in both the technological and human systems that were utilized in this response.

\section{Related Work}
This paper concentrates on documenting trends in the use of sUAS for hurricane response by emergency management agencies as opposed to the economic and humanitarian recovery phase (e.g., Typhoon Morakot (Taiwan, 2009)\cite{adams2011survey}).
sUAS have matured from the first use at the Hurricane Katrina response (2005)\cite{pratt2009conops}, followed by documented uses at Hurricanes Harvey (2017)\cite{fernandes2018quantitative}, Florence (2018)\cite{florence_use}, Michael (2018)\cite{fernandes2019quantitative}, and Ian (2022)\cite{manzini2023wireless}. The list is incomplete, as the Center for Robot-Assisted Search and Rescue (CRASAR) has flown for agencies having jurisdiction for four other hurricanes (Irma (2017), Laura (2020), Sally (2020), and Ida (2021)), but did not publish analyses; there may also be unreported instances from other events. 

This paper follows the analyses established for documenting the use of drones at Hurricanes Harvey \cite{fernandes2018quantitative} and Michael \cite{fernandes2019quantitative}: types of missions, models of drones used and why, types of data produced and volume, operations tempo, and human-robot interaction surprises. It extends the discussion to include failures and errors. Other published data on the use of sUAS during hurricane responses, such as Florence \cite{florence_use}, provide summaries but not details. 

\section{Uses of sUAS at Hurricane Ian}

The sUAS operations at Hurricane Ian began following the arrival of the storm on September 29, 2022 and continued for one week until October 5, 2022. sUAS operations were were exclusively conducted between the period of sunrise to sunset. This section describes the missions that were assigned to the different sUAS squads on each day of the response and recovery.

\begin{figure*}
    \centering
    \includegraphics[width=16cm]{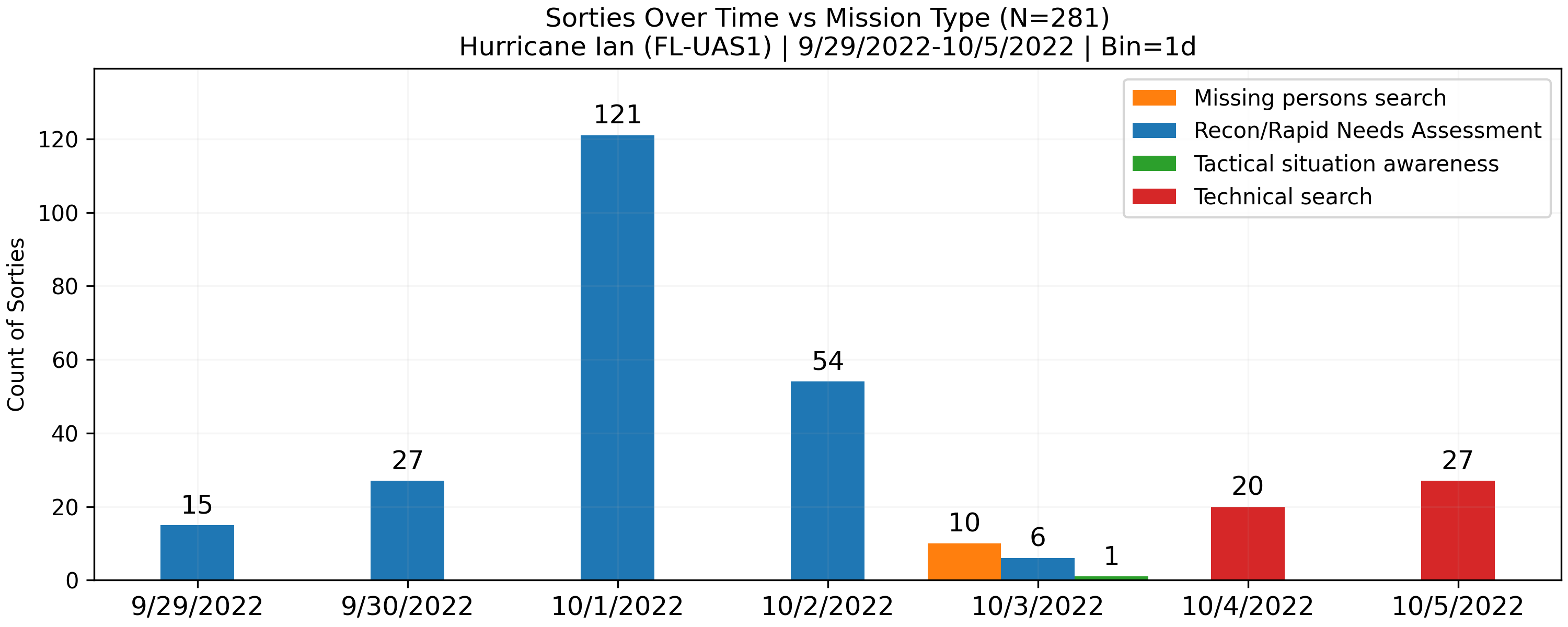}
    \caption{The count of sorties flown per day grouped by mission type.}
    \label{fig:sorties_mission_types}
\end{figure*}

\subsection{Mission Types}

Four types of missions were flown at Hurricane Ian as part of the formal response.

\begin{itemize}

\item \textbf{Recon/rapid needs assessment}: This type of mission focuses on using drones to rapidly estimate the damage to infrastructure (e.g., which neighborhoods are the most impacted? what are the access routes?) and conducting hasty searches for people who may be in distress. These missions are time-sensitive and typically flown first person view (FPV) and pilots generally take only a few representative images or short videos to illustrate the findings. In some cases, incident command will ask for a complete coverage survey of an area. In that case the imagery is either examined individually or stitched together as a low-resolution orthomosaic in the field rather than relying on uploading imagery to the cloud.  That same imagery may be used at a later date for stitching into a high-resolution orthomosaic, though it may be reflown at a lower altitude and higher percentage of image overlap to increase the resolution and quality of the resulting map. Recon/rapid needs assessment differs from flying for formal damage assessment or for detailed structural inspection, as these missions are generally directed by public works engineers and uses different procedures. This mission was previously referred to as Strategic Situation Awareness (SA), Survey, and Reconnaissance in \cite{fernandes2018quantitative, fernandes2019quantitative}; the change in terminology reflects a refinement of missions and skills.

\item \textbf{Tactical Situational Awareness}: In this type of mission, the drone squad provides tactical operational information to responders, such as a safety overwatch and a guide for responders to the target area. Drone squads directly communicate with the response teams and may provide streaming video.

\item \textbf{Missing persons search}: This type of search attempts to locate one or more specific missing persons within a large bounded area (e.g., persons who did not evacuate and presumed to be swept inland). It differs from recon/rapid needs assessment because it presumes the person is in distress. This mission was previously referred to as Ground Search in \cite{fernandes2018quantitative, fernandes2019quantitative} but was updated to the broader term for finding people in the water, trapped on boats, etc. 

\item \textbf{Technical search}: This type of search is conducted in the interior or at the threshold of a building, where GPS position is greatly diminished and the drone may experience significant wind shear. The intent is to search for victims or pets, capture the structural condition of the building for safe entry, and general documentation. The drone is out of the line of sight of the squad. This requires specialized piloting skills. It takes its name from  urban search and rescue technical searches through buildings.

\end{itemize}

\subsection{sUAS Operations}

FL-UAS1 conducted 281 sorties in support of 44 missions during Hurricane Ian.
A breakdown of these sorties and missions by mission type is included in Table \ref{tab:mission_types}. 
Missions were dispatched by request of the incident command. These missions fell into 5 of the universal set of 9 categories of disaster drone missions; four categories were not relevant for this type of disaster.
A summary of each day of operations is provided below.

\textbf{September 29, 2022}: On day one of the response, all of the missions performed large scale damage assessment and inspection of significant infrastructure, specifically bridges. All missions were conducted in the coastal area surrounding Fort Myers.
In total, 6 missions were conducted by 4 organizations flying 15 sorties.

\textbf{September 30, 2022}: The same profile continued into the second day of the response; however, on this day the team also conducted its first mapping mission of the Fort Myers Beach area. This mission returned data that was not immediately useful due to a software issue, which will be discussed later in Section \ref{sec:human_fails}. This day was also notable because the team sent an sUAS squad to central Florida to collect imagery of flooding that had been reported inland in the Wauchula area. 5 missions were conducted by 5 organizations flying 27 sorties.

\begin{table}[]
\centering
\begin{tabular}{|l|c|c|}
\hline
\textbf{Mission Type}          & \textbf{Sorties} & \textbf{Missions} \\ \hline
Recon/Rapid Needs Assessment   & 223              & 40                \\ \hline
Tactical Situation Awareness   & 1                & 1                 \\ \hline
Damage Assessment              & 0                & 0                 \\ \hline
Mapping and Geospatial Support & 0                & 0                 \\ \hline
Detailed structural inspection & 0                & 0                 \\ \hline
Wellness Checks                & 0                & 0                 \\ \hline
Missing persons search         & 10               & 1                 \\ \hline
Delivery                       & 0                & 0                 \\ \hline
Technical search               & 47               & 2                 \\ \hline
\textbf{Total}                 & \textbf{281}     & \textbf{44}       \\ \hline
\end{tabular}
\caption{Count of sorties and missions by mission type}
\label{tab:mission_types}
\end{table}

\textbf{October 1, 2022}: The third day of the response marked a significant change in operational scale and mission type. On this day, 10 organizations participated in 14 missions, flying 121 sorties. Due to the increase in number of available pilots and team members on this date, squads began performing dual-purpose missions involving ``Infrastructure Damage" assessment and ``Rapid Low Resolution Mapping" concurrently. 

\textbf{October 2, 2022}: The fourth day of the response saw a transition to almost exclusively mapping related missions. On this day, 10 organizations participated in 9 missions, flying 52 sorties. Dual-purpose missions involving ``Infrastructure Damage" assessment and ``Rapid Low Resolution Mapping" at the same time were flown. Only one mission on this day did not involve mapping.

\textbf{October 3, 2022}: sUAS operations on this date were centered around the San Carlos Island area, and focused on locating and searching boats that were adrift or pushed into the mangroves. First, sUAS teams were dispatched to locate boats throughout the area of operations (AO) to capture nadir imagery. This imagery was then processed to create a map of all located boats. Once all target boats were located, the teams were re-tasked to conduct low altitude primary searches of the vessels that were not easily accessible. 

While these missions were recorded separately, all sUAS activities on this date were tied to the same overall search mission.
Three missions were assigned that focused on searching for displaced boats potentially containing victims around San Carlos.
Three missions were assigned for ``Rapid Low Resolution Mapping" to generate a base map for subsequent ground and marine search operations.  
One mission focusing on ``Tactical situation awareness" was assigned and an sUAS squad was directed to provide overhead intelligence, surveillance, and reconnaissance (ISR) for ongoing operations in the San Carlos area.
Finally, following the completion of these above mission, 8 teams were assigned to a single mission focused on performing missing persons searches in the San Carlos area.
On this date, 8 organizations participated in 8 missions, flying 17 sorties.

\textbf{October 4, 2022}:
This day represented another change away from the types of missions that had been assigned to squads.
All squads were tasked with a single mission: flying technical interior searches of the structures on Fort Myers Beach. While substantial ground-based urban search and rescue (US\&R) resources were assigned to Fort Myers Beach, the scope and scale of the damage to structures made searches of elevated floors and interiors difficult and time consuming. To expedite searches of these areas, US\&R command tasked sUAS resources to assist in the search. This represents the first occurrence of this type of search operation during a large scale disaster.
On this date, 5 organizations participated in 1 mission, flying 20 sorties.

\begin{figure*}
    \centering
    \includegraphics[width=16cm]{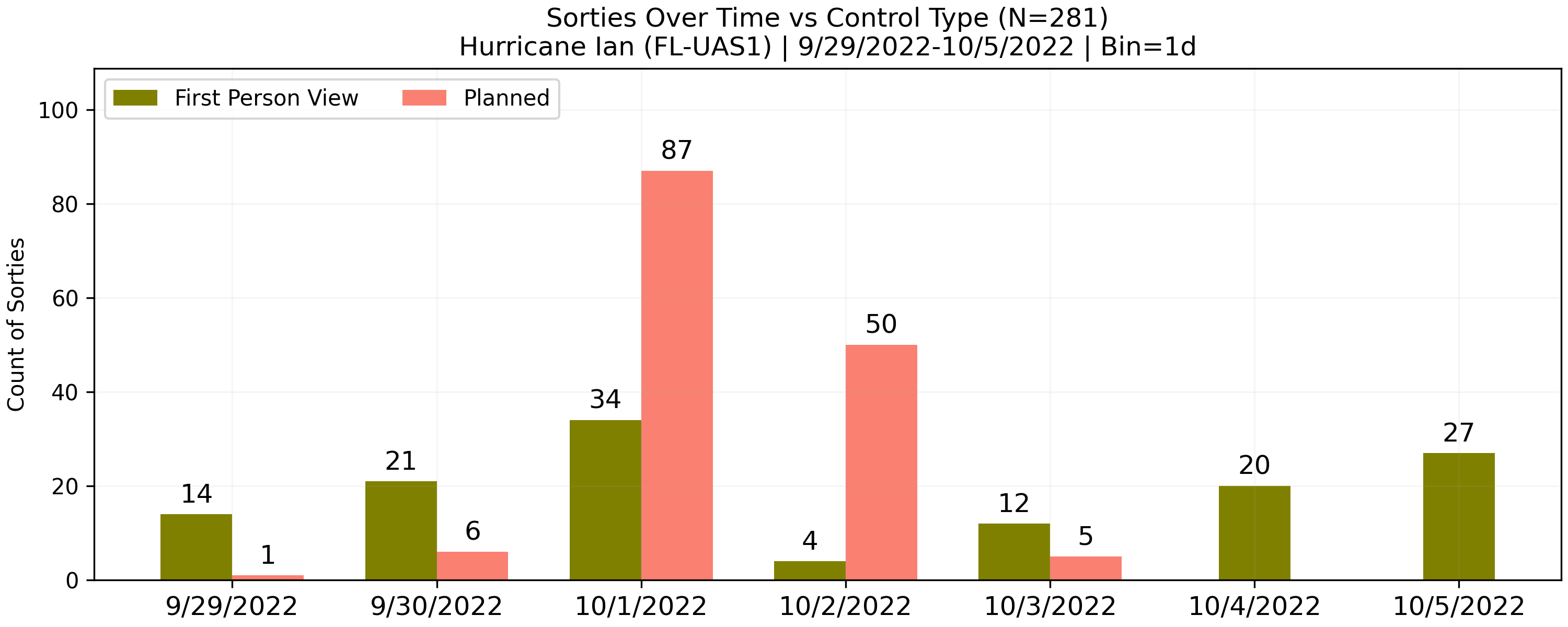}
    \caption{The count of sorties flown per day grouped by control type.}
    \label{fig:sorties_control_type}
\end{figure*}

\textbf{October 5, 2022}:
This day continued the effort of the prior day. Squads continued to fly technical, interior searches of the various structures on Fort Myers Beach. On October 5, squads were tasked to support US\&R operations on Fort Myers Beach. sUAS squads worked directly with US\&R squads as they continued primary and secondary searches of destroyed structures. sUAS squads also performed searches of partially submerged vehicles and boats in the interior canals of Fort Myers Beach. 
On this date, 5 organizations participated in 1 mission, flying 27 sorties.

A plot of the distribution of sorties compared to the different assigned mission types are provided in Figure \ref{fig:sorties_mission_types}.
This figure highlights how operational objectives shifted as the response progressed.
At the start of the response, missions were primarily focused on rapidly reacting to the ongoing needs of the response. 
However, following the first 72 hours (the nominal response period), an initial shift is observed where efforts focused instead on mapping. This was then followed by an additional shift as objectives change and squads are tasked to focus on technical search activities. 
This shift in objectives is also observed in changes in controls mode as shown in Figure \ref{fig:sorties_control_type}. 
Viewing this change in objectives via the control mode, the primary control mode initially starts in FPV as squads fly damage assessment and reconnaissance missions, the control mode switches to planned flying as objectives change to mapping, and then back to FPV as objectives again transition to technical search.

\subsection{Comparison to Prior Large Scale Use Of sUAS}
sUAS operations at Hurricane Ian were largely consistent with previous large scale sUAS disaster responses as squads routinely traveled at least an hour from the FOB to their mission destination, almost exclusively flew rotorcraft sUAS, and were expected to be available for tasking from sunrise to sunset.
The ``Recon/Rapid Needs Assessment" mission type dominated the first 72 hours, with a transition to other mission types afterwards \cite{fernandes2018quantitative, fernandes2019quantitative}. 
Finally, consistent with past large scale sUAS operations, lack of sufficient wireless connectivity was a pervasive issue \cite{manzini2023wireless}.

There are notable instances where operations diverged from past large scale sUAS operations. 
First, while technical search missions have occured during the recovery phase of large scale disasters \cite{pratt2009conops}, Hurricane Ian was the first time where an sUAS was used for technical search operations during the response phase. 
Second, while fixed wing sUAS have been utilized during the recovery phase of past large scale disasters \cite{fernandes2018quantitative}, Hurricane Ian saw them used during the response phase (though this also occurred during the unpublished response to Hurricane Ida). 
In addition, this was also the first use of a fixed wing VTOL sUAS during a response phase. This vehicle type was invaluable at collecting collecting imagery over long ranges despite limited available launch space.

\section{Team Composition}
This overall response team was composed of individuals from 12 different organizations identified in Section \ref{sec:intro}. When members of the team were deployed on missions to the field, they were selected primarily along organizational lines. This resulted in variable mission based squads with sizes that were based on the resources of the responding organization. In four cases, members from different organizations were partnered together due to differing levels of operational experience.

\section{sUAS Systems}
In the response to Hurricane Ian, the team used 10 different models of sUAS. Eight of these models were rotorcraft systems, one was a fixed wing vertical takeoff and landing system, and one was a fixed wing horizontal takeoff and landing system. Of the 281 sorties that were flown, rotorcraft systems conducted 271 (96.4\%), and specifically DJI Mavics conducted 176 (62.6\%). A detailed break down by type and by model can be found in Table \ref{tab:drone_models}.

\begin{figure*}
    \centering
    \includegraphics[width=16cm]{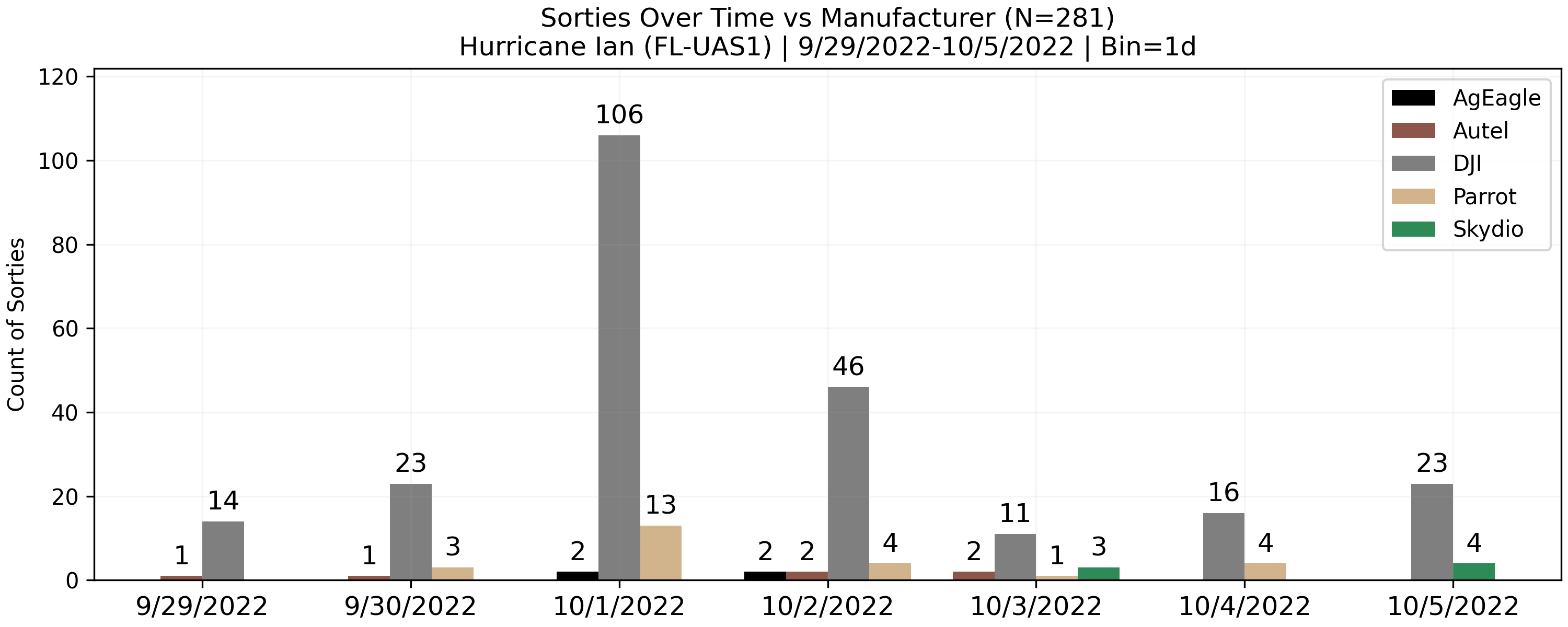}
    \caption{The count of sorties flown per day grouped by manufacturer.}
    \label{fig:sorties_manufacturer}
\end{figure*}

Viewing the same data in terms of sUAS manufactures, it is observed that DJI sUAS account for the vast majority of sorties; in fact 239 of the 281 sorties flown during Hurricane Ian 
 were flown by a DJI sUAS (85.1\%).
 This distribution is detailed in Figure \ref{fig:sorties_manufacturer}.

\begin{table}[]
\centering
\begin{tabular}{|cccc|}
\hline
\multicolumn{4}{|c|}{\textbf{Sorties By sUAS Type \& Model}}                                                         \\ \hline
\multicolumn{1}{|c|}{\textbf{sUAS Type}}      & \multicolumn{1}{c|}{\textbf{Sorties}}   & \multicolumn{1}{c|}{\textbf{sUAS Model}}          & \textbf{Sorties}   \\ \hline
\multicolumn{1}{|c|}{Fixed Wing}      & \multicolumn{1}{c|}{4}   & \multicolumn{1}{c|}{AgEagle EBEEX}          & 4   \\ \hline
\multicolumn{1}{|c|}{Fixed Wing VTOL} & \multicolumn{1}{c|}{6}   & \multicolumn{1}{c|}{Autel Dragonfish}       & 6   \\ \hline
\multicolumn{1}{|c|}{}                & \multicolumn{1}{c|}{}    & \multicolumn{1}{c|}{DJI Air2s}              & 2   \\ \cline{3-4} 
\multicolumn{1}{|c|}{}                & \multicolumn{1}{c|}{}    & \multicolumn{1}{c|}{DJI Matrice 210}        & 12  \\ \cline{3-4} 
\multicolumn{1}{|c|}{}                & \multicolumn{1}{c|}{}    & \multicolumn{1}{c|}{DJI Matrice 300}        & 13  \\ \cline{3-4} 
\multicolumn{1}{|c|}{Rotorcraft}                & \multicolumn{1}{c|}{271}    & \multicolumn{1}{c|}{DJI Matrice 30T}        & 36  \\ \cline{3-4} 
\multicolumn{1}{|c|}{}      & \multicolumn{1}{c|}{} & \multicolumn{1}{c|}{DJI Mavic 2 Enterprise} & 140 \\ \cline{3-4} 
\multicolumn{1}{|c|}{}                & \multicolumn{1}{c|}{}    & \multicolumn{1}{c|}{DJI Mavic Pro}          & 36  \\ \cline{3-4} 
\multicolumn{1}{|c|}{}                & \multicolumn{1}{c|}{}    & \multicolumn{1}{c|}{Parrot Anafi}           & 25  \\ \cline{3-4} 
\multicolumn{1}{|c|}{}                & \multicolumn{1}{c|}{}    & \multicolumn{1}{c|}{Skydio X2}              & 7   \\ \hline
\end{tabular}
\caption{Sorties per sUAS Type \& Model.}
\label{tab:drone_models}
\end{table}

\section{Analysis of Mission Sizes}
\begin{figure}
    \centering
    \includegraphics[width=9cm]{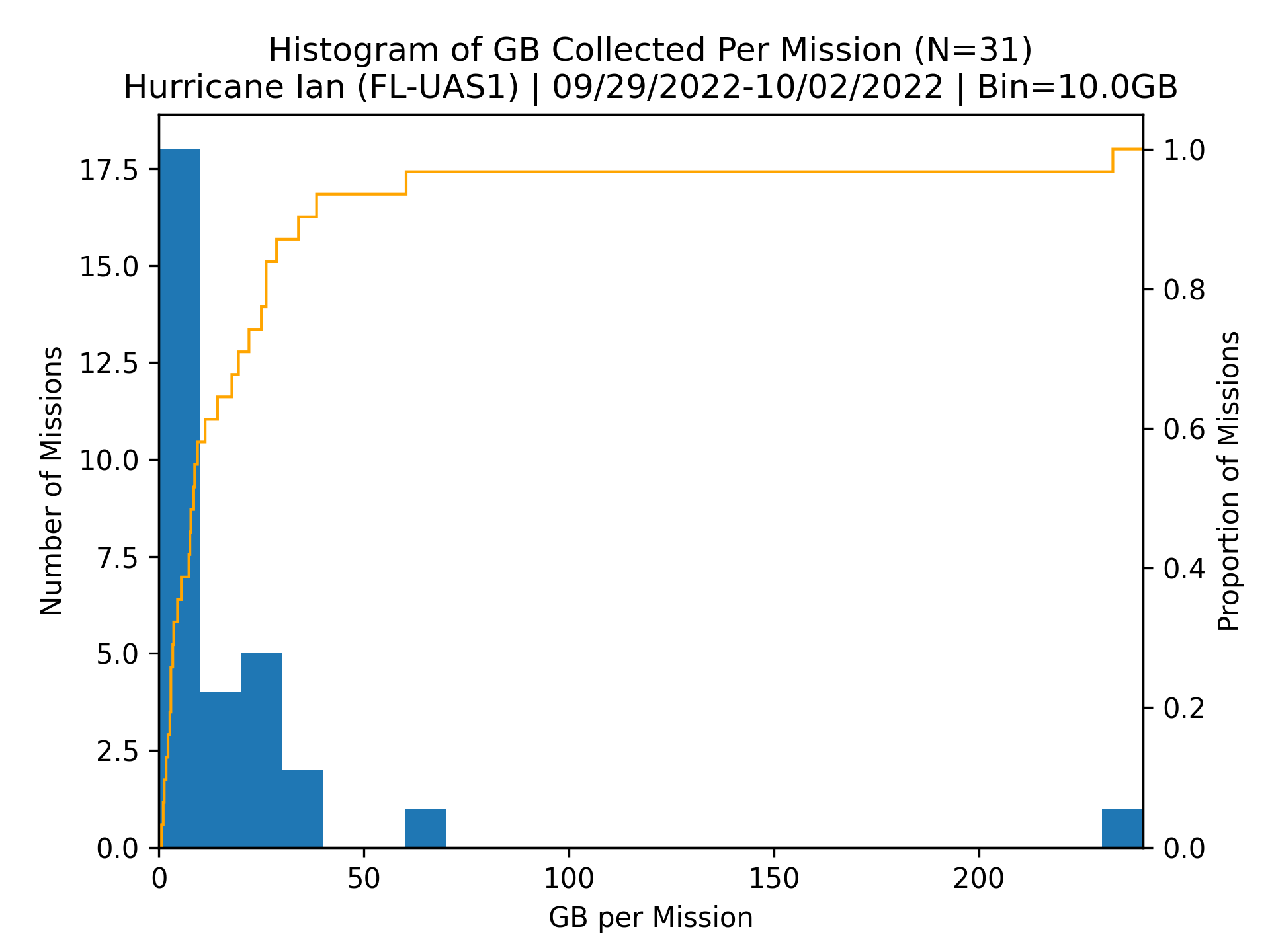}
\begin{table}[H]
\centering
    \setlength\tabcolsep{4pt}
\begin{tabular}{|l|ccccccc|}
\hline
\multicolumn{1}{|l|}{Percentile} & \multicolumn{1}{c|}{10th} & \multicolumn{1}{c|}{25th} & \multicolumn{1}{c|}{50th} & \multicolumn{1}{c|}{75th} & \multicolumn{1}{c|}{90th} & \multicolumn{1}{c|}{95th} & 99th    \\ \hline
Size \tiny{(N=31)} & 1.6\tiny{GB}                     & 2.9\tiny{GB}                     & 8.4\tiny{GB}                     & 24.8\tiny{GB}                    & 34.1\tiny{GB}                    & 60.2\tiny{GB}                    & 232.6\tiny{GB} \\ 
Files \tiny{(N=31)}                    & 2                         & 2                         & 164                       & 1329                      & 3360                      & 4711                      & 21632   \\ \hline
\end{tabular}
\end{table}
    \caption{Amount of data collected per mission.}
    \label{fig:mission_sizes}
\end{figure}

Further analysis of the data collected at Hurricane Ian was performed with the intention of developing guidance on the development of edge computing and data storage for sUAS operations in the future. Collected data from the period of September 29, 2022 to October 2, 2022 was analyzed. One finding of note is that the missions resulted in 8.5GB of data on average, and 60.2GB at the 95th percentile, which is shown in Figure \ref{fig:mission_sizes}. Two conclusions can be drawn from this information.

First, in terms of memory storage, missions can be reliably dispatched with blank 64GB memory cards, and memory cards with larger capacity and higher cost can be reserved for data intensive missions.

Second, when comparing this data with the average distance an sUAS team had to travel for their mission (p50=12mi, p95=35mi) \cite{manzini2023wireless}, in order to reliably outperform physically driving the data to its destination wireless connectivity should have an approximate capacity of 8GB/hr (17.8Mb/s), and ideally 60GB/hr (133.3Mb/s).


\section{Failures and Points of Friction}

During the sUAS response to Hurricane Ian, two pervasive points of friction appeared that were consistent with past responses: lack of sufficient wireless connectivity and coordination with manned aircraft. Both of these are discussed below, followed by a survey of failures encountered in the field, both technological and human.

\subsection{Operational Points of Friction}

Consistent with past sUAS operations at large scale disasters, the sUAS operations at Hurricane Ian suffered because of a lack of wireless and wired connectivity in the field.
As a result, squads were unable to transmit data collected in the field to cloud services for dissemination.
This problem has already been discussed in detail in \cite{manzini2023wireless}. 
As a result, data managers had to periodically physically transport hard drives of data to local high speed connection points at local universities and emergency operations centers (EOCs).

In addition, the sUAS squads frequently found themselves having to pause their missions due to manned aerial assets that were operating in the area.
This was partly mitigated by the issuance of a temporary flight restriction (TFR) on September 30 \cite{ian_notam}, but it remained an issue on the other days of the response. 
Other sUAS teams responding to large scale disasters are encouraged to track encounters with manned aircraft in order to further quantify, and develop mitigation strategies for this phenomenon.

\subsection{Technological Failures}
On the technological front, one software issue and two in flight failures occurred.

One drone squad encountered a software issue that resulted in their mapping missions being flown with an approximately 45 degree oblique camera angle instead of the nadir camera angle that was specified.
This behavior resulted in the collection of unusable data for two mapping missions. 
The software issue was found to be repeatable, and was unsuccessfully troubleshooted.
Following this the drone was removed from operational rotation.

FL-UAS1 also experienced the loss of two sUAS during operations. 
The first sUAS lost was an DJI Mavic Pro that lost power in flight and was not able to be recovered due to local conditions. At the time of the loss, the sUAS was performing a mapping mission on Fort Myers Beach. A post flight inspection was not performed and the data collected during this mission was lost.
The second sUAS lost was a Skydio X2E that collided with an overhead obstacle while performing indoor, technical search activities on Fort Myers Beach.
The sUAS and its data were recovered with minor physical damage and it was replaced by the manufacture the next day. 

\subsection{Human Operator Failures}
\label{sec:human_fails}
In terms of failures that were made by human operators, FL-UAS1 encountered two.
First, a single instance of an unauthorized flight occurred on October 1 where an sUAS squad flew a survey mission of the Fort Myers Pier. There was no assigned objective for this flight.
Imagery associated with this flight was handed off to data managers.

Second, a drone team was assigned to a residential area on a mission of type ``Recon/Rapid Needs Assessment" with an assigned objective of ``Infrastructure Damage."
While the team did perform this task, they also attempted to collect mapping imagery of the area, but were unfamiliar with the data collection process for mapping missions.
As a result, instead of collecting still imagery from a nadir view, the team hand flew video imagery using an unknown flight path. This resulted in unnecessary sUAS flight time and unusable data for mapping.
Imagery associated with this flight was handed off to data managers.

\section{Outcomes and Guidance for Future Responses}

Having now presented sUAS operations at Hurricane Ian, this section will present the potential directions and improvements that can be implemented at future large scale disaster sUAS operations. 

\subsection{Outcomes for sUAS Operations}
The coordination with manned aircraft remains a substantive point of friction in large scale sUAS operations. 
Future operations are encouraged to engage early with manned aircraft operations to coordinate operations and reduce conflicts.

For individual sUAS pilots, further emphasis should be placed on training for the common mission types described in Figure \ref{fig:sorties_mission_types}. Additionally, sUAS pilots should verify sensor settings, such as viewing angle, prior to any data capture.
Finally, sUAS pilots should be reminded to not perform unnecessary flights that do not further their assigned mission. These flights delay data transfer and incur additional risk.

\subsection{Outcomes for sUAS Data Management}
With the lack of wireless connectivity being such a substantial operational burden, future large scale sUAS operations are encouraged to establish partnerships with local EOCs as soon as possible in order to ensure effective wired connectivity in the absence of wireless connectivity.
Establishing these partnerships early will ensure that the periods of time where critical data must be communicated, will not be spent searching for adequate connectivity.

The data management workflow at Hurricane Ian involved collecting memory cards from sUAS squads and copying their contents to terabyte scale external storage devices that would then be physically transported to wired connections for uploading. 
While this approach has worked in the past, for future large scale sUAS operations, data managers are encouraged to utilize a network attached storage device for the storage and replication of data collected in the field. 
A network attached storage device would simplify this process by allowing multiple data managers to write concurrently to shared storage which could then be physically transported to wired connection points.

\section{Conclusion}
This document summarized the 281 sorties supporting the 44 missions conducted during the response and recovery to Hurricane Ian.
It provided analysis of the different types of missions flown, the different types of control modes used by sUAS pilots, and the different models, types, and manufactures of those sUAS.
This work reported on points of friction, and errors made in the field, both human and technological.
Finally, outcomes and future work was discussed for sUAS teams, data managers, and technological infrastructure.

\section*{Acknowledgment}

Acknowledgement and thanks is given to Justin Adams and all participants of FL-UAS1.
In addition acknowledgement and thanks to the IT department at the Collier County EOC who enabled the reliable transmission of FL-UAS1 data. This work was supported in part by the National Science Foundation under Grant No. CMMI-2306453

\bibliographystyle{IEEEtran}
\bibliography{main}

\end{document}